\documentclass{article}

\usepackage{microtype}
\usepackage{graphicx}
\usepackage{subcaption}
\usepackage{booktabs} 
\usepackage[utf8]{inputenc}
\usepackage[T1]{fontenc}
\usepackage{tcolorbox} 
\tcbuselibrary{skins}
\usepackage{amsmath}
\usepackage{amssymb}

\usepackage[preprint]{icml2026}

\usepackage{tabularx}   
\usepackage{ragged2e}   
\usepackage{bbm}
\usepackage{cuted}
\usepackage{caption}
\usepackage{multirow}
\usepackage{colortbl}
\usepackage{enumitem}

\definecolor{mycitecolor}{HTML}{4527A0}  
\definecolor{mylinkcolor}{HTML}{1565C0}   
\definecolor{myurlcolor}{HTML}{C62828}    

\usepackage{xcolor}
\usepackage[colorlinks=true,
            linkcolor=mylinkcolor,
            citecolor=mycitecolor,
            urlcolor=myurlcolor]{hyperref}
\usepackage{threeparttable}
\usepackage{xcolor}
\usepackage{tikz}
\usetikzlibrary{shapes.geometric} 

\definecolor{mainGreen}{named}{teal}
\definecolor{mainBlue}{named}{blue}

\newcommand{\bgGreen}[1]{%
    \begingroup%
    \setlength{\fboxsep}{1pt}
    \smash{\colorbox{mainGreen!15}{#1}}
    \vphantom{#1}
    \endgroup%
}

\newcommand{\bgBlue}[1]{%
    \begingroup%
    \setlength{\fboxsep}{1pt}%
    \smash{\colorbox{mainBlue!15}{#1}}%
    \vphantom{#1}%
    \endgroup%
}

\newcommand{\iconHexBlue}{%
    \smash{
        \tikz[baseline={([yshift=-0.6ex]current bounding box.center)}]
        \node[fill=mainBlue, regular polygon, regular polygon sides=6, inner sep=0pt, rotate=30, minimum size=1.6ex] {};%
    }%
}

\newcommand{\iconCircBlue}{%
    \smash{%
        \tikz[baseline={([yshift=-0.6ex]current bounding box.center)}]
        \node[fill=mainBlue, circle, inner sep=0pt, minimum size=1.6ex] {};%
    }%
}

\newcommand{\iconTriUpGreen}{%
    \smash{%
        \tikz[baseline={([yshift=-0.6ex]current bounding box.center)}]
        \node[fill=mainGreen, regular polygon, regular polygon sides=3, inner sep=0pt, minimum size=1.6ex] {};%
    }%
}

\newcommand{\iconTriDownGreen}{%
    \smash{%
        \tikz[baseline={([yshift=-0.6ex]current bounding box.center)}]
        \node[fill=mainGreen, regular polygon, regular polygon sides=3, rotate=180, inner sep=0pt, minimum size=1.6ex] {};%
    }%
}

\tcbuselibrary{breakable} 

\usepackage{amsmath}
\usepackage{amssymb}
\usepackage{mathtools}
\usepackage{amsthm}

\usepackage[capitalize,noabbrev]{cleveref}

\theoremstyle{plain}

\theoremstyle{definition}

\theoremstyle{remark}

\usepackage[textsize=tiny]{todonotes}

\icmltitlerunning{Modular Gradient Surgery}

\usepackage{calc}

\newcommand{\pngBaidu}{\includegraphics[height=3ex]{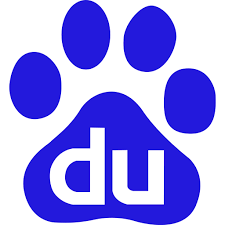}}
\newcommand{\pngUAlberta}{\includegraphics[height=3.3ex]{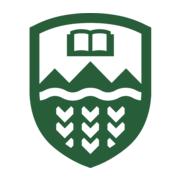}}

\icmlsetsymbol{baidu}{1}
\icmlsetsymbol{ualberta}{2}
\icmlsetsymbol{intern}{*}
\icmlsetsymbol{core}{$\dagger$}

\begin{document}

\twocolumn[
\icmltitle{
Advancing General-Purpose Reasoning Models with Modular Gradient Surgery 
}




\begin{icmlauthorlist}
\icmlauthor{Min Cai}{baidu,ualberta,intern}
\icmlauthor{Yu Liang}{baidu}
\icmlauthor{Longzheng Wang}{baidu}
\icmlauthor{Yan Wang}{baidu}
\icmlauthor{Yueyang Zhang}{baidu}
\icmlauthor{Long Xia}{baidu}
\icmlauthor{Zhiyuan Sun}{baidu}
\icmlauthor{Xi Ye}{ualberta}
\icmlauthor{Daiting Shi}{baidu}
\end{icmlauthorlist}

\icmlcorrespondingauthor{Min Cai, Xi Ye}{\{mcai8,xi.ye\}@ualberta.ca}
\icmlcorrespondingauthor{Daiting Shi}{shidaiting01@baidu.com}

\begin{center}
    \pngBaidu~\textsuperscript{1}\textbf{Baidu Inc.} \quad \pngUAlberta~\textsuperscript{2}\textbf{University of Alberta} \break
    \texttt{Code:\href{https://github.com/StringNLPLAB/MGS}{github.com/StringNLPLAB/MGS}}\break \texttt{Website:\href{https://modular-gradient-surgery.github.io}{modular-gradient-surgery.github.io}}
\end{center}
\vspace{-2pt}
\icmlkeywords{Machine Learning, ICML}

\vskip 0.3in

]

\newcommand{\cm}[1]{{\color{blue}[{CM: #1}]}}
\newcommand{\experiment}[1]{{\color{red}[{Experiment TODO: #1}]}}
\newcommand{\mathchattwostage}{\textsc{MCTwoStage~}}
\newcommand{\twostage}{\textsc{Sequential~}}
\newcommand{\singlestage}{\textsc{Mixed~}}
\newcommand{\Math}{\textit{Math}~}
\newcommand{\Chat}{\textit{Chat}~}
\newcommand{\Module}{\textsc{Module}~}
\newcommand{\Global}{\textsc{Global}~}

\printAffiliationsAndNotice{%
    \textsuperscript{*}Work done during CM's internship at Baidu
}

\begin{abstract}
Reinforcement learning (RL) has played a central role in recent advances in large reasoning models (LRMs), yielding strong gains in verifiable and open-ended reasoning. However, training a single general-purpose LRM across diverse domains remains challenging due to pronounced domain heterogeneity. Through a systematic study of two widely used strategies, \twostage RL and \singlestage RL, we find that both incur substantial cross-domain interference at the behavioral and gradient levels, resulting in limited overall gains.
To address these challenges, we introduce \textbf{M}odular \textbf{G}radient \textbf{S}urgery (\textbf{MGS}), which resolves gradient conflicts at the module level within the transformer. When applied to Llama and Qwen models, MGS achieves average improvements of 4.3 (16.6\%) and 4.5 (11.1\%) points, respectively, over standard multi-task RL across three representative domains (math, general chat, and instruction following). Further analysis demonstrates that MGS remains effective under prolonged training. Overall, our study clarifies the sources of interference in multi-domain RL and presents an effective solution for training general-purpose LRMs.

\end{abstract}
\begin{figure*}[t!]
    \centering
    \includegraphics[width=\linewidth]{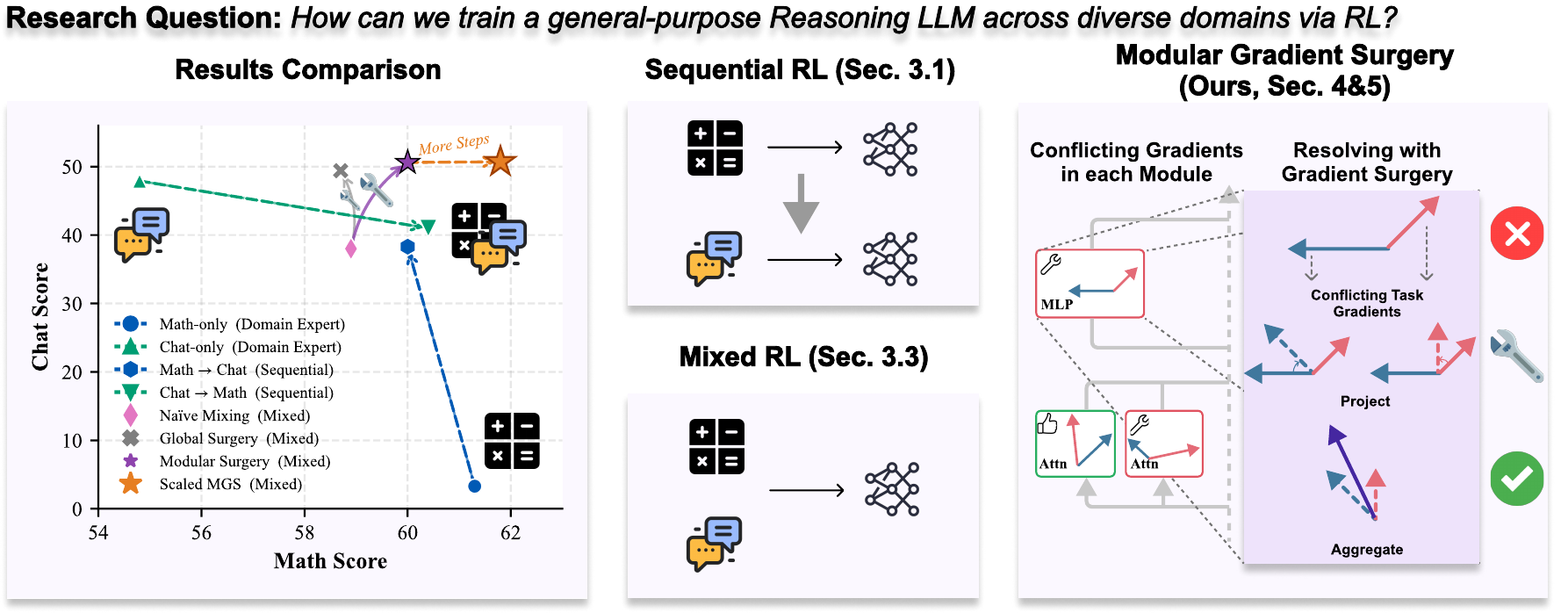}
    \caption{
    Effectiveness of different ways for training reasoning models on multiple domains. Naive strategies, such as sequential RL training (\twostage RL), or mixing different domains in the same batch (\singlestage RL), often result in limited performance across domains. We propose Modular Gradient Surgery (MGS), which resolves conflicting gradients at the module level and achieves the best multi-domain performance.
    }
    \label{fig:pipeline}
\end{figure*}
\section{Introduction}

Building large reasoning models (LRMs) with long chain-of-thought capabilities has become a central goal in recent language model development~\cite{liu2025deepseek,jaech2024openai,zhang2025survey}. Reinforcement learning (RL) has emerged as a key driver of this progress. In particular, reinforcement learning with verifiable rewards (RLVR) has demonstrated substantial gains in domains with objective correctness signals, such as mathematics~\cite{yu2025dapo,liu2025understanding} and code~\cite{wang2024enhancing,zhoubian2025rest}, where RL not only improves final accuracy but also incentivizes advanced reasoning such as branching and reflection~\cite{mondorf2024beyond,zeng2025simplerl,xu2025towards}. Motivated by these successes, recent efforts have extended RL beyond strictly verifiable settings, applying RL to more general tasks such as instruction following~\cite{gunjal2025rubrics,chang2025bleuberi} and open-ended chat~\cite{bhaskar2025language,li2025omni} using model-based rewards.

Despite this progress, training a single general-purpose reasoning model that performs well across diverse domains remains challenging. Building such a model requires combining heterogeneous rewards and domains (e.g., rule-based rewards for math, preference-based rewards for chat). Practitioners consistently observe difficulties when mixing these objectives: learning on one domain often degrades performance on another, transfer across domains is weak, and naively mixing rewards can lead to unstable or brittle training dynamics~\cite{li2025omni,bhaskar2025language,wang2025nemotron}.

In this work, we systematically study how to perform RL on diverse  domains to build broadly capable reasoning models. As illustrated in Figure~\ref{fig:pipeline}, we begin by analyzing two natural and widely used strategies: (1) \twostage RL, where a model is trained on different domains one after another; and (2) \singlestage (multi-task) RL, where data from multiple domains are combined within each training batch. Through controlled experiments on representative domains—math reasoning, general chat, and instruction following—we show that both strategies suffer from limitations in multi-domain settings. \twostage RL exhibits mode interference, where optimizing strongly for one domain erases or constrains behaviors needed for others, while \singlestage RL suffers from gradient conflicts between domains. As a result, both approaches exhibit a substantial gap compared to single-domain experts, as shown in Figure~\ref{fig:pipeline}.

To address these limitations, we propose \textbf{M}odular \textbf{G}radient \textbf{S}urgery (\textbf{MGS}), an effective method for mitigating cross-domain interference in \singlestage RL training. As pretrained transformers are highly modular, with different components specializing in distinct functionalities, and gradient conflicts are often localized to specific modules rather than uniformly distributed across the model. MGS resolves conflicting gradients at the module level using gradient surgery~\cite{yu2020gradient}, allowing compatible updates to be preserved while selectively mitigating destructive interference.

Empirically, we show that MGS consistently improves \singlestage RL training across model families and task combinations. Applied to Llama and Qwen models over \textit{Math}, \textit{Chat}, and Instruction-Following (\textit{IF}) domains, MGS substantially expands the capability frontier, achieving stronger and more balanced performance than \twostage training, Naïve Mixing, and Global Gradient Surgery. Further analysis demonstrates that MGS scales with extended training, yielding an additional 3\% relative improvement on the Math benchmark. Furthermore, MGS generalizes well to additional tasks, achieving a 19.4\% relative average improvement on \texttt{Llama-3.1-8B}. These gains are achieved with negligible overhead with high-performance parallel training frameworks like FSDP.

Overall, our findings clarify why multi-domain RL post-training remains difficult in practice and provide a principled approach for advancing general-purpose reasoning models. To summarize, our contributions include: 
(1) A systematic study of multi-domain RL for reasoning, revealing the limitations of \twostage and \singlestage training strategies;
(2) A new approach, Modular Gradient Surgery, which resolves cross-domain interference at modular levels; and
(3) An extensive evaluation across models and domains, establishing stong and consistent empirical gains of MGS.



\section{Background and Experimental Setup}
\label{sec:exp_setting}

\paragraph{Research question.} Our research goal focuses on how to effectively train a general-purpose reasoning model across diverse domains via RL. Previous works have shown strong success in training reasoning models with a single type of reward over a set of relatively similar domains. Notably, reinforcement learning with verifiable rewards (RLVR) trains reasoning models using rule-based rewards in domains such as mathematics and code~\cite{wen2025reinforcement,zeng2025simplerl,seed2025seed1,Guo2025DeepSeekR1}. Reinforcement learning with model-rewarded thinking (RLMT) operates with model-based rewards to optimize open-ended behaviors such as chat and creative writing~\cite{bhaskar2025language}. These approaches typically focus on one reward paradigm at a time. In contrast, our work studies RL training that jointly combines different prompts and rewards across domains.
 
 Formally, let $\pi_{\theta}$  denote reasoning LLM as a policy parameterized by $\theta$, which induces a conditional distribution $\pi(y,z\mid x)$ over responses $y$ and long chain-of-thought $z$ given a prompt $x$. 
 We consider a set of training domains denoted as $\mathcal{T} = \{(\mathcal{D}_k, R_k)\}_{k=1}^K$,  where each domain consists of a tuple of training data $D_{k}=\{x\}$, and a domain-specific reward function $R_{k}$. Given a prompt $x \sim \mathcal{D}_{k}$ and a response-cot pair $(y,z)$ sampled from the LLM policy $(y,z)\sim \pi_{\theta}(\cdot\mid x)$, a reward function produces a scalar reward $R(x, y) \in \mathbb{R}$.
Our goal is to optimize $\pi_{\theta}$ to perform well across these domains:
\begin{align}
    \max_\theta \; \mathbb{E}_{(\mathcal{D}_k,R_k)\sim\mathcal{T}} 
    \left[ \mathbb{E}_{x\sim\mathcal{D}_k,\, (y,z)\sim\pi_\theta(\cdot|x)} R_k(x,y) \right].
\end{align}

At evaluation time, we assess \textit{in-domain performance} with benchmarks from the training domains. In addition, we are interested in testing general reasoning capabilities beyond in-domain tasks with \textit{out-of-domain} benchmarks.

\subsection{Experimental Setup}
\begin{table}[ht]
\centering
\caption{Overview of Datasets and Prompt Types}
\label{tab:datasets}
\begin{tabularx}{\linewidth}{@{} l l >{\raggedright\arraybackslash}X @{}} 
\toprule
\textbf{Datasets} & \textbf{Size} & \textbf{Prompt Types} \\ 
\midrule

\texttt{SimpleRL-Zoo}   & 8,523 & \Math (GSM8K and MATH) \\
\texttt{Wildchat}        & 7,544 & \textit{Code, Creative writing, etc.} \\


\texttt{Nemotron} & 7,930 & \textit{Instruction Following} \\ 
\bottomrule
\end{tabularx}
\end{table}

\paragraph{Tasks and models.}
We use three representative domains for RL training, including 1) math reasoning (\textit{Math}), 2) general chat (\textit{Chat}), and 3) instruction following (\textit{IF}), and study two model families, \texttt{Qwen-2.5-7B}~\cite{yang2025qwen3} and \texttt{Llama-3.1-8B}~\cite{dubey2024llama}. Table~\ref{tab:datasets} summarizes the datasets used in experiments.
Specifically, for \textit{Math}, we use \texttt{SimpleRL-Hard} for \texttt{Qwen-2.5-7B} and \texttt{SimpleRL-Easy} for \texttt{Llama-3.1-8B}~\cite{zeng2025simplerl}, determined by model capability. Following RLVR~\cite{liu2025deepseek}, we adopt the correctness reward for math:
$R_{\textit{Math}} = \mathbbm{1}\{y=y_{ref}\}$, where $y_{ref}$ is the gold answer.
For \textit{Chat}, we follow \citet{bhaskar2025language} and use \texttt{Wildchat} as the training data. We provide scalar rewards $R_{\textit{Chat}}=r(x,y)$ where $r$ is the reward model. We use the open-source reward model \texttt{Skywork-Reward-V2-Llama-3.1-8B}~\cite{liu2025skywork}.
For \textit{IF}, we construct our training data (see Appendix~\ref{sec:data_construction}) based on \texttt{Nemotron}~\cite{liu2025prorl}, and use the scoring utilities in IFEval~\cite{zhou2023instruction} as our reward function: $R_{\textit{IF}}(x, y) = \frac{1}{|\mathcal{C}_x|} \sum_{c \in \mathcal{C}_x} \mathbbm{1}(y \models c)$, where $C_x$ is a list of constraints associated with input $x$.
These tasks, model families, and datasets are standard in recent RL-for-LLM literature~\cite{liu2025prorl,liu2025deepseek}.

\paragraph{RL Training.}
Across all experiments, we optimize the policy using a token-level GRPO objective used in DAPO~\cite{yu2025dapo}.
Formally, the objective is:
\begin{align}
J_{\text{DAPO}}(\theta)
=
\mathbb{E}_{\substack{(q,a)\sim\mathcal{D} \\ \{o_i\}_{i=1}^G \sim \pi_{\theta_{\mathrm{old}}}}}
\Bigg[
\frac{1}{\sum_{i=1}^{G}{|o_{i}|}}
\sum_{i=1}^G \sum_{t=1}^{|o_i|}
\nonumber \\
\min \Big(
r_{i,t}(\theta)\hat{A}_{i,t}, \text{clip}(r_{i,t}(\theta), 1-\epsilon, 1+ \epsilon)\hat{A}_{i,t}
\Big)
\Bigg],\label{eq:dapo_loss}\\
\text{where}~r_{i,t}(\theta)=\frac{\pi_\theta(o_{i,t}|q,o_{i,<t})}{\pi_{\theta_{\mathrm{old}}}(o_{i,t}|q,o_{i,<t})},\nonumber\\
\text{and}~\hat{A}_{i,t}=\frac{R_i-\text{mean}(\{R_i\}_{i=1}^{G})}{\text{std}(\{R_i\}_{i=1}^G)}.\nonumber
\end{align}
We also follow the hyper-parameters in previous literature~\cite{bhaskar2025language,yu2025dapo} if not explicitly mentioned. Concrete values for each hyper-parameter can be found in Appendix~\ref{appendix:hyper_param}.
\paragraph{Controlled Compute Budget.} To ensure a fair comparison and a controlled setting, we use a compute budget of 2 epochs for each domain across experiments in Sections 3--5, with the exception of the prolonged training experiments.
\paragraph{Evaluation Suites.}
We construct a comprehensive evaluation suite covering the three training domains, namely \textit{Math}, \textit{Chat}, and \textit{Instruction Following (IF)}. Our training datasets are drawn from and unify those used in prior RL-based post-training work for each domain~\cite{zeng2025simplerl,wang2025nemotron,bhaskar2025language}. In addition, we include a set of \textit{out-of-domain} benchmarks to assess generalization and overall reasoning capabilities. Specifically, we use:
\begin{itemize}[noitemsep,topsep=0pt,leftmargin=*]
    \item \textbf{Math}: MATH500~\cite{lightman2023let}, AIME24~\cite{aimo_validation_aime}, and GSM8K~\cite{cobbe2021training}
    \item  \textbf{Chat}: WildBench~\cite{lin2024wildbench}, AlpacaEvalv2~\cite{dubois2024length}, and CreativeWriting~\cite{paech2025eqbench}
    \item \textbf{I}nstruction \textbf{F}ollowing: IFBench~\cite{pyatkin2025generalizing} and IFEval~\cite{zhou2023instruction}
    \item \textbf{General} Capabilities: MMLU redux~\cite{gema2025we}, PopQA~\cite{mallen2023not}, and Zebra Logic~\cite{lin2025zebralogic}
\end{itemize}

Details of each benchmark are provided in Appendix~\ref{appendix:eval_suite}.

\section{Analyzing Cross-Domain Interference in Multi-Task RL}
In this section, we study and analyze cross-domain interference in \twostage and \singlestage RL.
\subsection{Experiment 1: \twostage RL}
Here, we demonstrate the limitations of \twostage RL for learning reasoning skills across multiple domains. Specifically, we show that \twostage RL suffers from cross-domain interference, which exhibits two failure modes:
(1) Forgetting: performance degradation on previously trained domains, and
(2) Rigidity: constrained learning on subsequently trained domains.
We collectively refer to this phenomenon as \textbf{Mode Interference}, reflecting competition between domain-specific optimization objectives.
\label{ref:exp_twostage}
\paragraph{Setting.} We study \twostage training on two domains: general chat ($\mathcal{D}_{\text{Chat}}$) and mathematical reasoning ($\mathcal{D}_{\text{Math}}$). We train
the model on one domain for 2 epochs, then continue RL training on the other domain for another 2 epochs, following the hyperparameters used in prior work~\cite{yu2025dapo,bhaskar2025language}. We consider both training orders: Chat$\rightarrow$Math and Math$\rightarrow$Chat.
All experiments use the token-level policy gradient loss defined in Eq.~(\ref{eq:dapo_loss}), with the domain-specific dataset $\mathcal{D}\in\{\mathcal{D}_{\text{Chat}},\mathcal{D}_{\text{Math}}\}$ applied in each stage. Importantly, the reference policy $\pi_{\text{ref}}$ is reset between stages, following standard practice~\cite{liu2025deepseek,wang2025nemotron}. All experiments are conducted on \texttt{Qwen-2.5-7B}.

\begin{figure}[t]
    \centering
    \includegraphics[width=\linewidth]{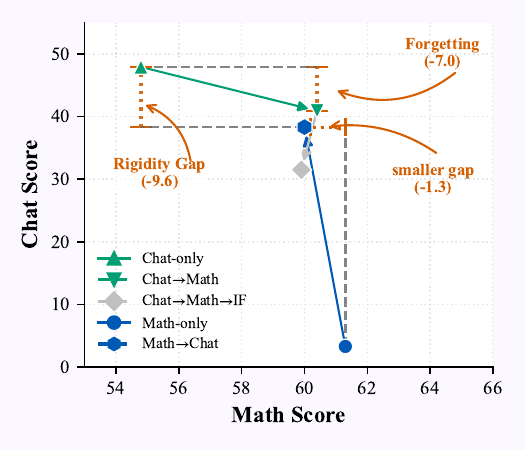}
    \caption{\textit{Chat} vs. \textit{Math} capabilities for \twostage RL training, with different training steps in the first stage. We identify Forgetting and Rigidity in this type of approach.}
    \label{fig:pareto_sequential}
\end{figure}

\paragraph{Results.}
Figure~\ref{fig:pareto_sequential} summarizes the performance of models trained under different \twostage configurations, compared against single-domain RL baselines and the base model. Full results are provided in Appendix~\ref{sec:add_seq_results}. Overall, Chat$\rightarrow$Math and Math$\rightarrow$Chat achieve similar trade-offs, with  Chat$\rightarrow$Math performing slightly better on average. 
Below, we highlight two key findings.

\paragraph{RL on one domain degrades performance on previously trained domains.}
RL training on a second domain \textbf{typically} reduces performance on the first domain, a phenomenon resembling \textbf{Catastrophic Forgetting}~\cite{luo2025empirical}.
While recent work suggests that RL is generally less prone to forgetting than supervised fine-tuning~\cite{chen2025retaining,chu2025sft}, our results show that it remains a significant limitation in sequential RL.
Crucially, however, we observe that this forgetting is \textbf{asymmetric} and significantly more severe when \Chat is the first stage.
As shown in Figure~\ref{fig:pareto_sequential}, \bgGreen{continuing RL on \Math after \Chat training} leads to a sharp drop in \Chat performance. \bgBlue{Training in the reverse order} results in less degradation in \Math capabilities. Additionally, it is also shown in the figure that continual RL training of Chat$\to$Math$\to$IF further downgrades the performance on both \Chat and \textit{Math}.

\paragraph{Furthermore, RL on one domain may rigidize training on subsequent domains.}

\begin{figure}[t]
    \centering
    \includegraphics[width=\linewidth]{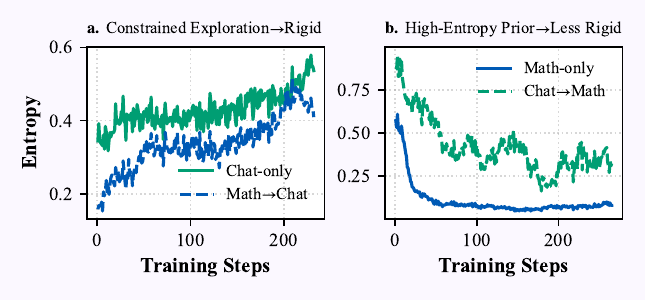}
    \caption{Entropy analysis on Reasoning Rigidity. \textbf{Left (Chat training)}: entropy dynamics comparison of  Math$\rightarrow$Chat and Chat-only. \textbf{Right (Math training)}: entropy dynamics comparison of Chat$\rightarrow$Math and Math-only.}
    \label{fig:entropy_analysis}
\end{figure}

Beyond forgetting, we observe a second and more subtle failure mode: training on one domain can constrain performance gains on subsequent domains, a phenomenon we define as \textbf{Reasoning Rigidity}. For example, Chat$\rightarrow$Math fails to reach the performance of $\text{Math-only}$ on \Math, and similarly Math$\rightarrow$Chat  underperforms Chat-only on \textit{Chat}.
We further observe that, the gap between \iconHexBlue~\bgBlue{Math$\rightarrow$Chat} and \iconTriUpGreen~\bgGreen{Chat-only} on \textit{Chat} (-9.6) is much larger than that between \iconTriDownGreen~\bgGreen{Chat$\rightarrow$Math} and \iconCircBlue~\bgBlue{Math-only} on \Math (-1.3).
This indicates an even subtler conclusion -- Reasoning Rigidity is more severe when \Math is trained first. This leads to a critical observation regarding RL training dynamics: the entropy usually decreases in \Math training and increases in \Chat training, and the entropy behavior affects subsequent training. As shown in Figure~\ref{fig:entropy_analysis} (left), when \Math is trained in the first stage, the overall entropy on \Chat data is lower than that of Chat-only training, which explains the Rigidity Gap in Figure~\ref{fig:pareto_sequential}. Conversely (right), \Chat training in the first stage brings a high-entropy prior to subsequent \Math training, which helps better exploration and explains the better \Math score in Chat$\rightarrow$Math.

\subsection{Analysis: Mode Interference in \twostage RL}

Our results in \S\ref{ref:exp_twostage} reveal the limitation of \twostage RL training, referred to as Mode Interference, which emphasizes that different domain objectives compete during RL optimization rather than composing additively.

\paragraph{Mode Interference limits the effectiveness of scaling \twostage training.} Increasing the amount of training on a single domain may not reliably translate into improved multi-domain performance, as gains in one stage can be offset by interference in previous or subsequent stages.

\paragraph{Properly ordering training domains mitigates mode interference.}
As shown in Figure~\ref{fig:pareto_sequential} and \ref{fig:entropy_analysis}, Chat$\to$Math generally has better performance. These results suggest that, in sequential RL, domains that preserve higher entropy and flexible reasoning behaviors should be trained earlier, as they form a more permissive foundation for later, more structured domains.

\subsection{Experiment 2: \singlestage RL}
We also evaluate \singlestage RL training. Unlike \twostage RL, which optimizes domains sequentially, \singlestage RL mixes data from multiple domains within each batch, allowing gradients from different domains to be applied simultaneously at every optimization step.
\paragraph{Setting.} Our primary setting uses a 1:1 mixture of general chat ($\mathcal{D}_{\text{Chat}}$) and mathematical reasoning ($\mathcal{D}_{\text{Math}}$) prompts. We keep the batch size identical to that used in \twostage training and train for 2 epochs over the mixed dataset. As a result, the total number of optimization steps matches that of \twostage training, ensuring a fair comparison. 
Unlike \twostage training, the reference policy $\pi_{\text{ref}}$ remains fixed as the initial Base model throughout training, following standard practice in multi-task RL~\cite{li2025omni,zeng2025zero}. In addition to the balanced mixture, we evaluate different mixing ratios, including 1:9 and 9:1 for Math versus Chat, to study how different mixing portions affect cross-domain trade-offs.

\begin{figure}[t]
    \centering
    \includegraphics[width=\linewidth]{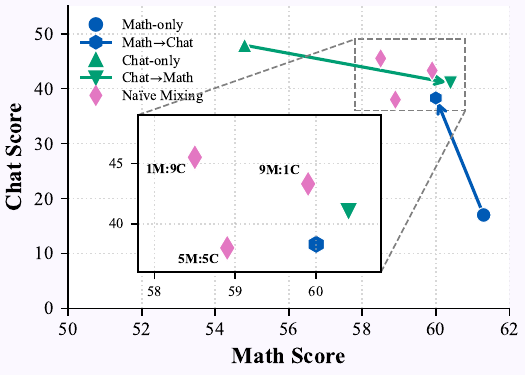}
    \caption{\singlestage RL training with different data proportions on \texttt{Qwen-2.5-7B}. \Math performance is positively correlated with the number of \Math data, while the correlation on \Chat is more complex. Compared to \twostage RL, \singlestage RL checkpoints generally have  higher \Chat scores with lower \Math scores.}
    \label{fig:pareto_normal_mixing}
\end{figure}

\paragraph{Results.} 
Figure~\ref{fig:pareto_normal_mixing} provides a detailed summary of the performance of \singlestage RL under different Math–Chat mixing ratios. Similar to \twostage training, we observe persistent negative interplay between domains: models trained with mixed-domain batches consistently underperform the best domain-specific experts.
Increasing the proportion of Math data leads to monotonic improvements in Math performance. However, even with a highly skewed mixture (90\% Math, 10\% Chat), the resulting model still lags behind a Math-only model trained for 2 epochs. This suggests that simply adjusting data proportions is insufficient to eliminate cross-domain interference in \singlestage RL.

\subsection{Analysis: Gradient conflicts in \singlestage RL}
\begin{figure}[t]
    \centering
    \includegraphics[width=\linewidth]{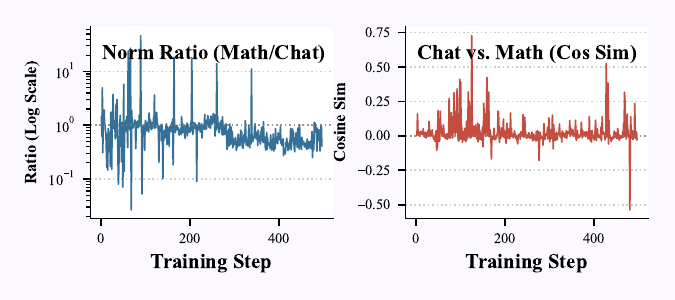}
    \caption{Gradient cosine similarity between \Math, \Chat in \texttt{Llama-3.1-8B} during training. Gradient norm ratio between \Math and \Chat in is also shown. Gradient conflict and imbalance exist between different tasks across different training steps.}
    \label{fig:grad_conflict}
\end{figure}
To better understand the source of negative cross-domain interactions in \singlestage RL, we analyze the gradients obtained from different domains, which is a commonly used diagnostic for multi-domain training~\cite{yu2020gradient}. Specifically, we examine the alignment between gradients computed from Chat and Math samples across training steps.

As shown in Figure~\ref{fig:grad_conflict}, gradient conflicts frequently arise during training, indicating that updates that improve performance on one domain can directly interfere with optimization on the other. These conflicts limit the effectiveness of naive multi-domain RL training. 
These observations motivate the need for methods that explicitly detect and resolve cross-domain gradient conflicts. In the next section (\S\ref{sec:mgs}), we introduce Modular Gradient Surgery, an effective approach for mitigating these conflicts during multi-domain RL.

\section{Modular Gradient Surgery}
\label{sec:mgs}
To mitigate gradient conflicts in \singlestage training, we introduce a new technique that resolves interference at the module level. Our approach addresses gradient conflicts within individual transformer modules by applying gradient surgery~\cite{yu2020gradient} locally to each module. We present our method in the following paragraphs.
\label{sec:method_mgs}
\paragraph{Background: Gradient Surgery.} Negative interference between tasks has been a long-standing challenge in multi-task learning. A representative line of research attributes this to conflicting task gradients, leading to the development of gradient surgery methods. These methods address conflicting objectives in multi-task optimization by modifying gradients to avoid destructive interference. A representative approach is PCGrad~\cite{yu2020gradient}, which detects conflicts between task gradients (e.g., via negative cosine similarity) and projects gradients to remove components that harm other tasks.
Suppose there are two tasks, let $\mathbf{g}_i = \nabla_\theta \mathcal{L}_i$ and $\mathbf{g}_j = \nabla_\theta \mathcal{L}_j$ denote the gradients for tasks $i$ and $j$. PCGrad identifies a conflict when the inner product is negative, i.e., $\mathbf{g}_i \cdot \mathbf{g}_j < 0$. To resolve this, $\mathbf{g}_i$ is replaced by its projection onto the orthogonal complement of $\mathbf{g}_j$, and vice versa:
\begin{align}
    \mathbf{g}_i^{\text{proj}} &= \mathbf{g}_i - \frac{\mathbf{g}_i \cdot \mathbf{g}_j}{\| \mathbf{g}_j \|^2} \mathbf{g}_j \nonumber\\
    \mathbf{g}_j^{\text{proj}} &= \mathbf{g}_j - \frac{\mathbf{g}_j \cdot \mathbf{g}_i}{\| \mathbf{g}_i \|^2} \mathbf{g}_i \nonumber\\
    \mathbf{g}^{\text{new}} &= \mathbf{g}_i^{\text{proj}} + \mathbf{g}_j^{\text{proj}}.
\end{align}
Here, $\mathbf{g}^{\text{new}}$ is the final gradient used for optimization.
\paragraph{Modular gradient surgery (MGS).}

\definecolor{global_color}{RGB}{0, 0, 180}   
\definecolor{module_color}{RGB}{180, 80, 0}  

\begin{algorithm}[ht]
   \caption{Modular Gradient Surgery}
   \label{alg:mgs}
\begin{algorithmic}[1]
   \REQUIRE Parameters $\theta$, Batch $\mathcal{B}$ for tasks $\{1 \dots K\}$
   
   \STATE \textbf{Definitions:} 
   \STATE $\quad$ Let $\theta$ be partitioned into disjoint modules $\Phi$ s.t. $\theta = \bigcup_{\phi \in \Phi} \phi$.
   \STATE $\quad$ Let $\text{Flatten}(\mathbf{g})$ map a gradient tensor (or list of tensors) to $\mathbb{R}^D$.
   \STATE $\quad$ Let $\text{Unflatten}(\mathbf{v})$ map a vector $\mathbf{v} \in \mathbb{R}^D$ back to the shape of $\mathbf{g}$.
   
   \STATE \textcolor{gray}{// 1. Compute initial gradients}
   \STATE $\mathbf{g}_k \leftarrow \sum_{\tau \in \mathcal{B}_k} \nabla_\theta \mathcal{L}(\tau; \theta) \quad \forall k \in \{1 \dots K\}$
   \STATE $\mathbf{g}_k^{\text{PC}} \leftarrow \mathbf{g}_k \quad \forall k$ \quad \textcolor{gray}{// Initialize working gradients}
   
   \STATE \textcolor{gray}{// 2. Apply Gradient Surgery}
   \FOR{task $i$ in $\{1 \dots K\}$}
      \FOR{task $j$ in $\text{RandomPermutation}(\{1 \dots K\} \setminus \{i\})$}
            \STATE \textcolor{gray}{// Apply PCGrad independently per module}
            \FOR{each module $\phi \in \Phi$}
                \STATE \textcolor{module_color}{Let $\mathbf{v}_i, \mathbf{v}_j = \text{Flatten}(\mathbf{g}_i^{\text{PC}}[\phi]), \text{Flatten}(\mathbf{g}_j[\phi])$}
                \IF{$\mathbf{v}_i \cdot \mathbf{v}_j < 0$}
                    \STATE $\mathbf{v}_i \leftarrow \mathbf{v}_i - \frac{\mathbf{v}_i \cdot \mathbf{v}_j}{\|\mathbf{v}_j\|^2 + \epsilon} \mathbf{v}_j$
                    \STATE \textcolor{module_color}{$\mathbf{g}_i^{\text{PC}}[\phi] \leftarrow \text{Unflatten}(\mathbf{v}_i)$}
                \ENDIF
            \ENDFOR

      \ENDFOR
   \ENDFOR
   \STATE \textbf{return} $\Delta \theta = \sum_i \mathbf{g}_i^{\text{PC}}$
\end{algorithmic}
\end{algorithm}
We propose Modular Gradient Surgery (MGS), which extends the PCGrad framework by resolving gradient conflicts at the module level. Our approach is motivated by the observation that pretrained transformers are inherently modular, with distinct components specializing in different functional domains. Prior work shows that MLP layers are closely associated with parametric memory and knowledge storage~\cite{meng2022locating,elhage2021mathematical,du2025post}, while attention layers primarily handle information routing and reasoning~\cite{li2023inference,todd2023function,zhang2025query,wu2024retrieval,olsson2022context}, with further specialization across depth~\cite{artzy2024attend}.
This modularity implies that gradient conflicts are not uniform across the network. In practice, we observe substantial variation in gradient conflict frequency and magnitude across modules during multi-domain RL (Figure~\ref{fig:grad_conflict} and \ref{fig:grad_conflict_modular}).
Some modules may experience severe cross-domain interference, while others exhibit largely compatible gradients.
This further motivates module-aware conflict resolution, whereas applying global gradient surgery can therefore be overly conservative.
Algorithm~\ref{alg:mgs} details the MGS method. As shown in the algorithm, MGS partitions model parameters into modules (e.g., attention and MLP blocks by layer in \texttt{model.named\_parameters}) and applies gradient surgery independently within each module.

\definecolor{graybg}{RGB}{245,245,245}
\definecolor{headerbg}{RGB}{230,230,240}

\begin{table*}[ht]
\centering
\caption{\textbf{Main Results on different benchmarks.} Comparison of post-training strategies. We report the average for Math (M. Avg) and Chat (C. Avg) separately. The highest scores are in bold, and the second best are underlined.}
\label{tab:main_results}

\setlength{\tabcolsep}{3.2pt} 
\renewcommand{\arraystretch}{1.2}

\begin{tabular}{ll cccc cccccc c}
\toprule

& & \multicolumn{4}{c}{\textbf{Math (pass@1, avg@4)}} & \multicolumn{4}{c}{\textbf{Chat}} & \multicolumn{2}{c}{\textbf{Misc.}}\\
\cmidrule(lr){3-6} \cmidrule(lr){7-10} \cmidrule(lr){11-12}

\rowcolor{headerbg}
\textbf{Backbone} & \textbf{Method} & \textbf{MATH} & \textbf{GSM} & \textbf{AIME} & \textbf{M. Avg.} & \textbf{WB} & \textbf{AE2} & \textbf{CW3} & \textbf{C. Avg.} & \textbf{IF} & \textbf{Gen.} & \textbf{Total Avg.} \\
\midrule

\multicolumn{13}{c}{\cellcolor{graybg}\textit{Prior Works}} \\
\midrule

\multirow{1}{*}{Llama}
 & RLMT & 4.9 & 10.2 & 0.0 & 5.0 & 35.6 & 15.6 & 36.3 & 29.2 & 20.4 & 21.5 & 18.9 \\
\arrayrulecolor{black!10}\midrule\arrayrulecolor{black}

\multirow{2}{*}{Qwen}
 & RLMT & 71.8 & 82.7 & 10.0 & 54.8 & 44.7 & 57.2 & 41.7 & 47.9 & 25.9 & 35.4 & 42.4 \\
 & General Reasoner & 69.4 & 81.5 & 16.7 & 55.9 & 13.3 & 38.8 & 39.2 & 30.4 & 25.1 & 36.7 & 38.1 \\
\arrayrulecolor{black!10}\midrule\arrayrulecolor{black}

\multirow{1}{*}{Qwen-Inst}
 & OpenThinker v3 & 78.8 & 87.7 & 29.2 & 65.2 & -50.2 & 19.3 & 29.7 & -0.4 & 24.4 & 39.1 & 32.8 \\
\midrule

\multicolumn{13}{c}{\cellcolor{graybg}\textit{Multi-task Baselines}} \\
\midrule

\multirow{2}{*}{Qwen}
 & Normalized Adv. & 76.1 & 91.2 & 9.2 & 58.8 & 43.8 & 44.8 & 39.0 & 42.5 & 27.3 & 34.2 & 41.9 \\
 & Model Merging & 74.8 & 90.6 & 11.7 & 59.6 & 33.0 & 12.4 & -24.8 & 6.9 & 24.8 & 32.9 & 31.4 \\
\midrule

\multicolumn{13}{c}{\cellcolor{graybg}\textbf{\textit{This Work}}} \\
\midrule

\multirow{5}{*}{Llama}
 & Math$\to$Chat & 24.6 & 71.6 & 0.0 & 32.1 & \textbf{25.7} & 29.0 & \textbf{38.3} & \underline{31.0} & 21.4 & 25.0 & 27.9 \\
 & Chat$\to$Math & \underline{27.6} & \underline{78.8} & \textbf{1.7} & \underline{36.0} & -35.0 & 6.5 & 28.3 & -0.1 & 21.3 & 28.8 & 21.5 \\
 & Naïve Mixing & 25.8 & 76.6 & 0.0 & 34.1 & 15.0 & 21.5 & 33.8 & 23.4 & 20.7 & 23.7 & 25.9 \\
 & Global Surgery & \textbf{29.8} & \textbf{80.4} & \textbf{1.7} & \textbf{37.3} & 2.3 & \textbf{48.3} & \underline{34.9} & 28.5 & \underline{22.9} & \textbf{29.4} & \underline{30.1} \\
 & \textbf{Modular Surgery} & 26.0 & 73.6 & \underline{0.8} & 33.5 & \underline{24.7} & \underline{39.6} & 34.1 & \textbf{32.8} & \textbf{23.0} & \underline{29.3} & \textbf{30.2} \\
\arrayrulecolor{black!10}\midrule\arrayrulecolor{black}

\multirow{6}{*}{Qwen}
 & Math$\to$Chat & 75.9 & \underline{90.8} & \underline{13.3} & \underline{60.0} & 35.7 & 35.4 & \underline{43.9} & 38.3 & 24.8 & 30.8 & 39.7 \\
 & Chat$\to$Math & \textbf{78.2} & \textbf{92.2} & 10.8 & \textbf{60.4} & 36.0 & 46.8 & 40.1 & 41.0 & 24.4 & \underline{36.5} & 42.0 \\
 & Naïve Mixing & \underline{76.0} & 89.8 & 10.8 & 58.9 & \underline{47.1} & 26.2 & 40.7 & 38.0 & 22.1 & \underline{36.5} & 40.4 \\
 & Global Surgery & 73.0 & 88.8 & \textbf{14.2} & 58.7 & 45.4 & \textbf{62.5} & 40.2 & \underline{49.4} & \underline{25.6} & \textbf{36.8} & \underline{44.2} \\
 & \textbf{Modular Surgery} & 75.1 & 90.6 & \textbf{14.2} & \underline{60.0} & \textbf{48.3} & \underline{59.3} & \textbf{44.2} & \textbf{50.6} & \textbf{28.2} & 35.3 & \textbf{44.9} \\

\bottomrule
\end{tabular}
\end{table*}

\section{Advancing General-Purpose Reasoning with Modular Gradient Surgery}
In this section, we demonstrate the effectiveness of MGS in \S\ref{sec:results_mgs}, and provide extended experiments in \S\ref{sec:extended_exp}.
\subsection{Experiment 3: \singlestage Training with MGS}
\label{sec:results_mgs}
\paragraph{Baselines.}
We compare our methods with several multi-task baselines:
1) Model Merging~\cite{ablin2025soupofexperts,wortsman2022model}, we use a simple average method as our baseline.
2) Normalized Advantage~\cite{zhang2025agentrl}, this is a multi-task baseline that normalizes token-level advantages of the same tasks to reduce inter-task variance.
In addition, we include the results of Global Gradient Surgery (GGS) for comparison. For the details of other baselines, please refer to Appendix~\ref{sec:details_single_task_baselines}.
\paragraph{Result Comparisons.}
As shown in Table~\ref{tab:main_results}, the results show that MGS achieves the best performance over Naïve Mixing, \twostage RL, and other baselines. Specifically, MGS consistently achieves the best performance on both \texttt{Qwen} and \texttt{Llama}, with absolute improvements of 4.3 and 4.5 in total averaged scores over Naïve Mixing. In the meantime, \twostage RL exhibits significant performance trade-offs. For instance, the Chat$\rightarrow$Math approach on \texttt{Llama} results in a Chat Averaged performance of -0.1. In contrast, MGS effectively harmonizes conflicting tasks, achieving the highest \textit{IF} scores and \textit{Chat} scores across both backbones and consistently outperforming other multi-task baselines, e.g., Normalized Advantage and Model Merging.

\paragraph{MGS outperforms previous general models.} As also shown in Table~\ref{tab:main_results}, MGS achieves better performance compared to previous general models, i.e., RLMT and General Reasoner in \textit{Math}, \textit{Chat}, and \textit{IF}.

\paragraph{GGS might be overly conservative.} GGS is worse than MGS, and even in some cases, Naïve Mixing. This may be due to the overly conservative nature of GGS. For example, a conflict in a single, less-critical module could force GGS to project the entire model's gradient, potentially stalling learning in modules where there was no conflict at all.

\subsection{Experiment 4: Scaling Training Steps and Domains}
\label{sec:extended_exp}
We have showcased the effectiveness of MGS trained on \textit{Math} and \textit{Chat} under a controlled training budget. In this section, we further demonstrate that MGS helps advance general-purpose reasoning (\S\ref{subsec:advancing_results}) by scaling our setting to (1) training on three tasks and (2) training with more steps. In addition, we provide an analysis of MGS from a mechanistic viewpoint in \S\ref{subsec:mgs_analysis_module_norm}, and an analysis of MGS overhead can be found in Appendix~\ref{sec:additional_mgs}.
\paragraph{Setting.} To demonstrate that resolving gradient conflicts also helps \singlestage training with more than two tasks, we blend in the third task, i.e., instruction following (\textit{IF}). We also conduct a prolonged training experiment to demonstrate that MGS scales with more training steps. All settings remain the same as the previous ones.

\paragraph{Additional Baselines.}
For prolonged training, we compare MGS with \twostage RL, and for mixing three tasks, we compare MGS with Naïve Mixing and GGS.

\label{subsec:advancing_results}
\paragraph{MGS consistently surpasses other baselines as the number of tasks increases.}











\begin{table}[ht]
\centering
\caption{\texttt{Llama-3.1-8B} results of mixing \textit{Math}, \textit{Chat}, and \textit{IF}.}
\label{tab:math_chat_if_shorten}

\renewcommand{\arraystretch}{1.1}

\begin{tabular}{l ccccc}
\toprule
\rowcolor{headerbg}
\textbf{Method} & \textbf{Math} & \textbf{Chat} & \textbf{IF} & \textbf{Gen} & \textbf{Avg} \\
\midrule
Naïve Mixing  & 35.0 & 22.9 & 25.4 & 25.2 & 27.3 \\
Global Surgery & \underline{35.8} & \underline{24.2} & \textbf{31.1} & \underline{30.3} & \underline{30.3} \\
\textbf{Mod. Surgery} & \textbf{36.0} & \textbf{30.5} & \underline{30.0} & \textbf{33.1} & \textbf{32.6} \\ 

\bottomrule
\end{tabular}
\end{table}
We show results of three task setting in Table~\ref{tab:math_chat_if_shorten} (full results see Table~\ref{tab:math_chat_if}, Appendix~\ref{sec:additional_mgs}). As shown in the table for results on \texttt{Llama-3.1-8B}, MGS still achieves the best averaged performance over Naïve Mixing and GGS. MGS exhibits a more balanced behavior, with the highest scores on Math, Chat, and General QAs. As shown in Table~\ref{tab:math_chat_if}, MGS with three tasks also achieves the highest averaged performance on both \texttt{Llama-3.1-8B} and \texttt{Qwen-2.5-7B}.

\paragraph{RL with MGS scales with increased training steps.}
\begin{figure}[t]
    \centering
    \includegraphics[width=\linewidth]{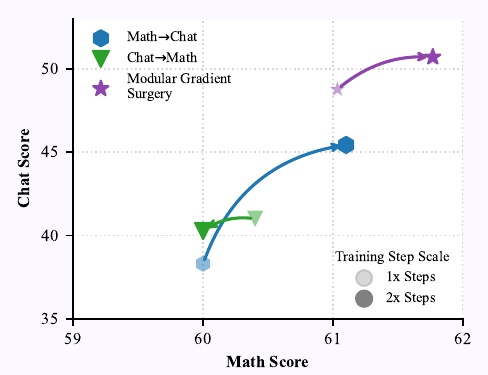}
    \caption{Performance trend comparison as training steps scaled up. Training steps are doubled against the normal setting.}
    \label{fig:scale_training_steps}
\end{figure}
In the normal setting, each model is trained with 2 epochs, which is around 480 steps. As shown in Figure~\ref{fig:scale_training_steps} of the Appendix~\ref{sec:additional_mgs}, we present the performance with a doubled number of training steps. Overall, MGS achieves better performance and scales as the number of training steps increases. Although Math$\rightarrow$Chat also gets better scores with more training steps, its performance is far behind MGS. In comparison, Chat$\rightarrow$Math degrades in both axes, which reveals the risks of Mode Interference in \twostage RL.

\subsection{Analysis}
\label{subsec:mgs_analysis_module_norm}
\begin{figure}[t]
    \centering
    \includegraphics[width=\linewidth]{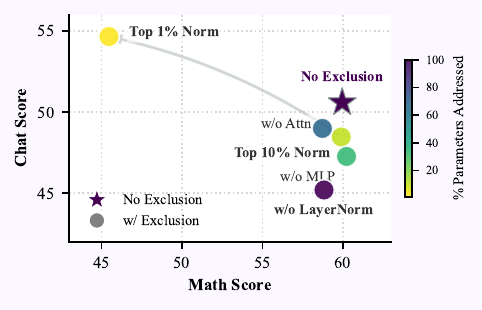}
    \caption{Ablation study of MGS module selection. (1) Ablation by module type (\texttt{Attn}, \texttt{MLP}, and \texttt{LayerNorm}). (2) Ablation by gradient norm ranking (Top-$\mu\%$). \%Parameters Addressed refers to the proportion of model parameters to which MGS is applied.}
    \label{fig:ablation_modules_norm}
\end{figure}

In this section, we provide: (1) an evaluation of how gradient norm magnitude correlates with these contributions, and (2) a mechanistic analysis of gradient dynamics to identify which model components contribute most to cross-task interference.

\paragraph{Effect of gradient norm.}
We conduct another ablation study based on gradient norms. In this experiment, we apply MGS solely to the modules whose gradient norms fall within the range of top-$\mu\%$ ($\mu\in\{1, 10\}$). Because for each module, we have $K$ task gradients, we use the maximum norm as the candidate to compare with other modules:
\begin{align}
    \|\mathbf{g}[\phi]\|_{\text{candidate}} = \max_{k \in \{1, \dots, K\}} \|\mathbf{g}_k\|_2,
\end{align}
where $\mathbf{g}[\phi]$ refers to the gradients w.r.t parameter $\phi$.
Results are also shown in Figure~\ref{fig:ablation_modules_norm}, where we find that while applying MGS to modules of Top 10\% Norm exhibits a similar trend to excluding different modules, applying MGS to modules of Top 1\% Norms shows a distinct trade-off in model capability. With Top 1\% Norm, the model has a performance boost in \Chat, but at the same time, \Math performance drops, with only marginal improvements compared to the Base model. This suggests that while high-norm modules are critical for \textit{Chat}, \Math relies on the coordinated alignment of gradients across a broader range of modules, not just those with the largest magnitude. This observation also suggests that gradient norm magnitude may serve as an effective heuristic for identifying target modules for gradient surgery and warrants further investigation.

\paragraph{Effect of different modules.}
We observe that the cosine similarity between the tasks has different patterns across different modules (Fig.~\ref{fig:grad_conflict_modular}). Therefore, we try to answer the question of \textit{``How addressing gradient conflict of certain modules contributes to the final performance''}. Specifically, we focus on \texttt{MLP}, \texttt{Attention Head} (\texttt{Attn}), and we also consider \texttt{LayerNorm} as a type of transformer module. We selectively exclude gradient surgery for specific module families in this experiment. For example, as shown in Figure~\ref{fig:ablation_modules_norm}, ``w/o MLP'' means MGS is applied to everything \textit{except} MLP. As shown in the Figure, excluding different modules harms \textit{Chat} capability, while excluding \texttt{Attn} and \texttt{LayerNorm} harms \textit{Math} capability.
Among these, although \texttt{LayerNorm} contains the fewest parameters, excluding \texttt{LayerNorm} from MGS leads to a dramatic performance drop in both axes. Unlike \texttt{Attn} or \texttt{MLP} layers that encode semantic functions, \texttt{LayerNorm} regulates the feature statistics of the embedding space. This may indicate that conflicts in \texttt{LayerNorm} are significant and more likely to be pronounced in model behaviors.

\section{Related Work}
Our work is closely related to LLM post-training and multi-task RL.
LLM post-training has evolved from standard RLHF for alignment \citep{ouyang2022training} to Reinforcement Learning with Verifiable Rewards (RLVR) for complex reasoning \citep{shi2025deep,Guo2025DeepSeekR1}. While multi-task RL is well-studied in classical contexts, focusing on gradient manipulation \citep{yu2020gradient} and weighting \citep{chen2018gradnorm} to mitigate task interference, its application to LLM RL post-training remains nascent~\cite{wu2025imbalanced}. Recent efforts in multi-domain LLM RL \citep{li2025omni,liu2025deepseek} often rely on complex pipelines or distillation, frequently overlooking the inherent gradient imbalances identified in mixed-objective training \citep{wu2025imbalanced}. Our work bridges this gap by introducing Modular Gradient Surgery, an approach that resolves inter-task conflicts at the module level, providing a stable and efficient recipe for general-purpose RL. Our complete related works are shown in Appendix~\ref{sec:complete_related_works}.

\section{Conclusions}
In this paper, we conduct a systematic study on how to perform RL
on diverse domains to build broadly capable reasoning models. We study two basic training strategies, i.e., \twostage training and \singlestage mixed training, and find that 1) for \twostage RL, Mode Interference manifests as Forgetting and Rigidity;  2) for \singlestage RL, we observe gradient conflicts which pose negative interference on different tasks. To advance general-purpose reasoning, we propose Modular Gradient Surgery, which addresses the conflicts at the module level. Empirical results show that MGS consistently leads to better performance.

In addition, our work is the first to apply gradient manipulation in LLM RL post-training. 
We hope our work can serve as an initiative toward more effective and efficient LLM Post-training for general-purpose reasoning.

\section*{Impact Statement}
This work aims at studying and advancing general-purpose reasoning LLM, which mainly focuses on Math, Chat and Instruction Following. The deployment of such models may have broader social impacts, while none of which we feel must be specifically highlighted here.
\section*{Acknowledgment}
We would like to thank Bale Chen, Quang Hieu Pham, Siyuan Dai, Zhiyue Wu, Yichi Wang, Hanyu Lin, Yiqiang Ye, and Xuefeng Wu for their helpful comments, feedback, and discussions.
\bibliography{ref}
\bibliographystyle{icml2026}

\clearpage
\appendix
\onecolumn
\section{Details of Evaluation Suites}
\label{appendix:eval_suite}

In this section, we present the details of the evaluation suites employed in our study:

\begin{itemize}
    \item \textbf{MATH500}~\cite{lightman2023let}: A subset of the MATH benchmark consisting of 500 challenging, competition-level mathematics problems.
    
    \item \textbf{GSM8K}~\cite{cobbe2021training}: A benchmark comprising 8.5k high-quality grade school math word problems designed to test multi-step mathematical reasoning.
    
    \item \textbf{AIME24}~\cite{aimo_validation_aime}: A dataset containing problems from the 2024 American Invitational Mathematics Examination, evaluating advanced problem-solving capabilities.

    \item \textbf{WildBench}~\cite{lin2024wildbench}: This suite evaluates the model's ability to converse with users via 1,024 prompts, including multi-turn interactions. Unlike AlpacaEval 2, WildBench uses an instance-wise manual rubric to mitigate reward hacking. Responses are compared against a GPT-4 reference and scored as -100 (much worse), -50 (worse), 0 (similar), 50 (better), or 100 (much better). The final metric is the mean of these instance-wise scores.

    \item \textbf{AlpacaEval V2}~\cite{dubois2024length}: Contains 805 user prompts paired with reference responses from GPT-4-1106-preview. It outputs a head-to-head win rate (0\%--100\%) using a generative judge. We use the length-controlled win rate as recommended by~\citeauthor{dubois2024length}, substituting the default judge with GPT-4o.

    \item \textbf{Creative Writing V3}~\cite{paech2025eqbench}: Evaluates the ability to write 96 story chapters under various constraints. We generate an absolute score between 0--100 using Claude-Sonnet-4 as the judge.

    \item \textbf{IFEval}~\cite{zhou2023instruction}: Provides models with 541 simple questions under verifiable constraints (e.g., ``do not use commas''). It generates a score between 0--100 signifying how well the model adheres to instructions.

    \item \textbf{IFBench}~\cite{pyatkin2025generalizing}: Provides 294 prompts from varied domains, each containing multiple constraints (e.g., ``include three numbers in your 22nd sentence''). We average the compliance rate across all examples to produce a score between 0--100.

    \item \textbf{MMLU$_{\text{redux}}$}~\cite{gema2025we}: A manually cleaned version of the MMLU (Hendrycks et al., 2021) benchmark. It consists of 5,700 questions testing general knowledge across 57 subjects.

    \item \textbf{PopQA}~\cite{mallen2023not}: Consists of approximately 14k factual questions regarding popular and long-tail entities. The result is reported as a percentage score between 0\%--100\%.

    \item \textbf{ZebraLogic}~\cite{lin2025zebralogic}: Tests language models on 1,000 logical grid puzzles. We report the exact match accuracy ranging from 0\%--100\%.
\end{itemize}
\clearpage
\section{Data Construction for \textit{IF} Data}
\label{sec:data_construction}
The original \texttt{Nemotron} Instruction Following dataset contains 56,339 data points. To ensure that the number of data is close to that of \textit{Math} and \textit{Chat}, we uniformly sample 7,930 data according to the number of constraints $|C_x|$.

\section{Hyperparameters and Configurations}
\label{appendix:hyper_param}
In Table~\ref{tab:hyperparams_all}, we present the hyper-parameters used in \twostage and \singlestage training.
\begin{table}[t]
    \centering
    \begin{threeparttable}
        \caption{Hyperparameters for \twostage (\textit{Math}, \textit{Chat}, and \textit{IF}), and \singlestage training configurations.}
        \label{tab:hyperparams_all}
        
        \begin{tabular}{lcccc}
            \toprule
            & \multicolumn{3}{c}{\textsc{Sequential}} & \\
            \cmidrule(lr){2-4}
            Hyperparameter & \textit{Math} & \textit{Chat} & \textit{IF} & \textsc{Mixed} \\
            \midrule
            Batch size & 64 & 64 & 64 & 64 \\
            Samples per prompt & 8 & 8 & 8 & 8 \\
            Group size & 8 & 8 & 8 & 8 \\
            Max prompt length & 1024 & 896 & 896 & 896 \\
            Max response length & 4096 & 4096 & 4096 & 4096 \\
            Num. epochs & 2 & 2 & 2 & 2 \\
            Actor learning rate & 1e-6 & 1e-6 & 1e-6 & 1e-6 \\
            Weight decay & 0.01 & 0.01 & 0.01 & 0.01 \\
            Scheduler & constant & constant & constant & constant \\
            Warmup ratio & 0 & 0 & 0 & 0 \\
            Advantage estimator & GRPO & GRPO & GRPO & GRPO \\
            KL coefficient & 0.0001/0\tnote{d} & 0.001 & 0.001 & 0.001 \\
            Entropy coefficient & 0.001/0\tnote{d} & 0 & 0 & 0 \\
            \bottomrule
        \end{tabular}
        
        \begin{tablenotes}
            \small
            \item[d] We used 0 for both KL and Entropy coefficient at the same time following~\citeauthor{yu2025dapo}.
        \end{tablenotes}
    \end{threeparttable}
\end{table}
\begin{figure}[h]
    \centering
    \includegraphics[width=0.5\linewidth]{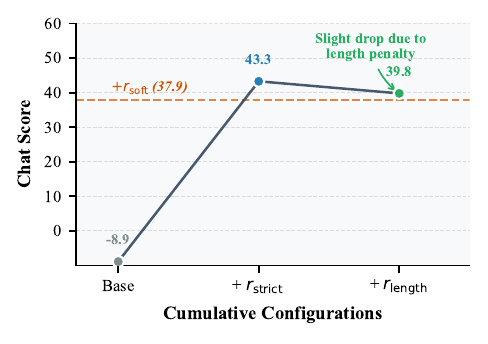}
    \caption{Configuration study of \Math training in Math$\to$Chat. Scores of the average of AlpacaEvalV2, WildBench, and CreativeWritingV3 when cumulatively adding different training configurations for the first stage \Math training. All the checkpoints evaluated here are trained with \Math and \Chat. The dashed-line shows the score of using soft format reward compared to the strict format reward.}
    \label{fig:math_then_chat_config}
\end{figure}
\begin{figure}
    \centering
    \includegraphics[width=0.5\linewidth]{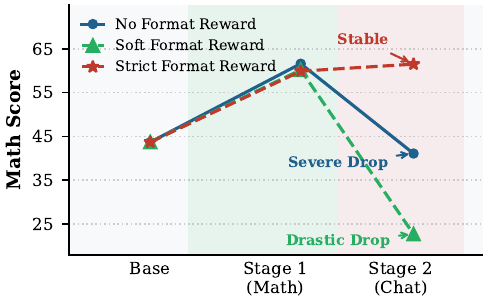}
    \caption{Results on \Math using different format rewards for \Math training in \twostage RL. No format reward or using soft format reward leads to performance drop, while with strict format reward, \twostage shows a stabler behavior.}
    \label{fig:format_study}
\end{figure}
Furthermore, we investigate how using different format reward and adding length penalty affects training. We adopt the length penalty in \citeauthor{yu2025dapo} shown below:
\begin{equation}
R_{\text{length}}(y) = 
\begin{cases} 
0, & |y| \leq L_{\text{max}} - L_{\text{cache}} \\
\frac{(L_{\text{max}} - L_{\text{cache}}) - |y|}{L_{\text{cache}}}, & \text{otherwise} \\
-1, & |y| > L_{\text{max}}
\end{cases}
\end{equation}
Figure~\ref{fig:math_then_chat_config} shows that adding a length penalty slightly downgrades the capability compared to simply using strict format reward. In addition, we also show that using soft format reward will also lead to lower \Chat score compared to using strict format reward. In our experiments, we by default adding length penalty to \singlestage to prevent unexpected model behaviors.
For format reward, we compare two variants: (1) soft format reward and (2) strict format reward.
As shown in Figure~\ref{fig:format_study}, training solely with correctness reward, the model achieves high performance on \Math in the first stage. However, in the second stage of \Chat training, the performance drops drastically. After adding a strict format reward, although the performance is lower, it continues to grow in the second stage, showing a stabler behavior. In addition, compared to adding soft format reward, we found that only after adding a strict format reward that requires models to generate ``\texttt{<think></think><response></response>}'' pairs in order, can we achieve stable \Math performance at the second stage. The strict vs. soft format rewards are:
\begin{align}
    &S_{\text{tags}} = (\texttt{<think>}, \texttt{</think>}, \texttt{<response>}, \texttt{</response>}) \\
    &R_{\text{soft}}(y) = \mathbbm{1}[ S_{\text{tags}} \subseteq y ] \\
    &R_{\text{strict}}(y) = \mathbbm{1}[ y \text{ follows order of } S_{\text{tags}} ].
\end{align}
The final math reward is multiplied by the format reward. For example, when using $R_{\text{strict}}$ as the format reward, the math reward $R_{\text{Math}}(x, y)$ is defined as
\begin{align}
    R^{\text{format}}_{\text{Math}}(x, y)=\mathbbm{1}[\text{correct}] * \mathbbm{1}[ y \text{ follows order of } S_{\text{tags}} ]
\end{align}
It is worth noting that we only used non-zero entropy coefficient in our preliminary study for ``No-Format Math'' and ``Soft Format Math'' as shown in Figure~\ref{fig:format_study}. And in our normal setting, we only use length penalty for \textit{Math} training.

\paragraph{Objective with KL divergence and entropy constraints.}
In our preliminary study, we explored an objective function for \textit{Math} tasks that incorporates both Kullback-Leibler (KL) divergence and entropy regularization to maintain policy stability and exploration:
\begin{align}
J_{\textsc{Sequential}}(\theta)
=
\mathbb{E}_{\substack{(q,a)\sim\mathcal{D} \\ \{o_i\}_{i=1}^G \sim \pi_{\theta_{\mathrm{old}}}}}
\Bigg[
\frac{1}{\sum_{i=1}^{G}{|o_{i}|}}
\sum_{i=1}^G \sum_{t=1}^{|o_i|}
\Big(
\min \Big(
r_{i,t}(\theta)\hat{A}_{i,t},
\operatorname{clip}(r_{i,t}(\theta), 1-\epsilon, 1+\epsilon)\hat{A}_{i,t}
\Big) \nonumber \\
- \beta\, \mathbb{D}_{\text{KL}}\!\big(\pi_\theta \parallel \pi_{\text{ref}}\big)_{i,t}
+ \gamma\, \operatorname{H}\!\big(\pi_\theta\big)_{i,t}
\Big)
\Bigg],\label{eq:two_stage}
\end{align}
where the token-level KL divergence and entropy are defined as:
\begin{equation}
\mathbb{D}_{\text{KL}}(\cdot)_{i,t}=\sum_{v \in \mathcal{V}} \pi_{\theta}(v|\cdot)
\log \frac{\pi_{\theta}(v|\cdot)}{\pi_{\text{ref}}(v|\cdot)},\text{~and~}
\operatorname{H}(\pi_\theta)_{i,t}=-\sum_{v \in \mathcal{V}} \pi_{\theta}(v|\cdot)\log \pi_{\theta}(v|\cdot).
\end{equation}

\paragraph{Policy entropy analysis.} To study the entropy dynamics during training, we measure the average token-level policy entropy across the training dataset $\mathcal{D}$. For a policy $\pi_{\theta}$, the mean entropy $\mathcal{H}$ is defined as:
\begin{equation}
\mathcal{H}(\pi_\theta, \mathcal{D}) = \mathbb{E}_{x \sim \mathcal{D}} \left[ \frac{1}{|\boldsymbol{y}|} \sum_{t=1}^{|\boldsymbol{y}|} \operatorname{H}(\pi_\theta)_{t} \right] = -\frac{1}{|\mathcal{D}|} \sum_{x \in \mathcal{D}} \frac{1}{|\boldsymbol{y}|} \sum_{t=1}^{|\boldsymbol{y}|} \sum_{v \in \mathcal{V}} \pi_{\theta}(v|\cdot) \log \pi_{\theta}(v|\cdot).
\end{equation}
In our experiments, we compute this metric over each batch of prompts used for training to monitor the evolution of policy diversity during training.
\clearpage
\section{Full Algorithm of Modular Gradient Surgery}
\label{sec:full_algorithm}

\definecolor{global_color}{RGB}{0, 0, 180}   
\definecolor{module_color}{RGB}{180, 80, 0}  

\begin{algorithm}[ht]
   \caption{Global vs. Modular Gradient Surgery}
   \label{alg:pcgrad}
\begin{algorithmic}[1]
   \REQUIRE Parameters $\theta$, Batch $\mathcal{B}$ for tasks $\{1 \dots K\}$, Mode $\in \{\textsc{Global}, \textsc{Module}\}$
   
   \STATE \textbf{Definitions:} 
   \STATE $\quad$ Let $\theta$ be partitioned into disjoint modules $\Phi$ s.t. $\theta = \bigcup_{\phi \in \Phi} \phi$.
   \STATE $\quad$ Let $\text{Flatten}(\mathbf{g})$ map a gradient tensor (or list of tensors) to $\mathbb{R}^D$.
   \STATE $\quad$ Let $\text{Unflatten}(\mathbf{v})$ map a vector $\mathbf{v} \in \mathbb{R}^D$ back to the shape of $\mathbf{g}$.
   
   \STATE \textcolor{gray}{// 1. Compute initial gradients}
   \STATE $\mathbf{g}_k \leftarrow \sum_{\tau \in \mathcal{B}_k} \nabla_\theta \mathcal{L}(\tau; \theta) \quad \forall k \in \{1 \dots K\}$
   \STATE $\mathbf{g}_k^{\text{PC}} \leftarrow \mathbf{g}_k \quad \forall k$ \quad \textcolor{gray}{// Initialize working gradients}
   
   \STATE \textcolor{gray}{// 2. Apply Gradient Surgery}
   \FOR{task $i$ in $\{1 \dots K\}$}
      \FOR{task $j$ in $\text{RandomPermutation}(\{1 \dots K\} \setminus \{i\})$}
         
         \IF{Mode is \textsc{Global}}
            \STATE \textcolor{gray}{// Flatten to perform projection in vector space}
            \STATE \textcolor{global_color}{Let $\mathbf{v}_i, \mathbf{v}_j = \text{Flatten}(\mathbf{g}_i^{\text{PC}}), \text{Flatten}(\mathbf{g}_j)$}
            \IF{$\mathbf{v}_i \cdot \mathbf{v}_j < 0$}
                \STATE \textcolor{gray}{// Project $\mathbf{v}_i$ onto orthogonal complement of $\mathbf{v}_j$}
                \STATE $\mathbf{v}_i \leftarrow \mathbf{v}_i - \frac{\mathbf{v}_i \cdot \mathbf{v}_j}{\|\mathbf{v}_j\|^2 + \epsilon} \mathbf{v}_j$
                \STATE \textcolor{global_color}{$\mathbf{g}_i^{\text{PC}} \leftarrow \text{Unflatten}(\mathbf{v}_i)$}
            \ENDIF
            
         \ELSIF{Mode is \textsc{Module}}
            \STATE \textcolor{gray}{// Apply PCGrad independently per module}
            \FOR{each module $\phi \in \Phi$}
                \STATE \textcolor{module_color}{Let $\mathbf{v}_i, \mathbf{v}_j = \text{Flatten}(\mathbf{g}_i^{\text{PC}}[\phi]), \text{Flatten}(\mathbf{g}_j[\phi])$}
                \IF{$\mathbf{v}_i \cdot \mathbf{v}_j < 0$}
                    \STATE $\mathbf{v}_i \leftarrow \mathbf{v}_i - \frac{\mathbf{v}_i \cdot \mathbf{v}_j}{\|\mathbf{v}_j\|^2 + \epsilon} \mathbf{v}_j$
                    \STATE \textcolor{module_color}{$\mathbf{g}_i^{\text{PC}}[\phi] \leftarrow \text{Unflatten}(\mathbf{v}_i)$}
                \ENDIF
            \ENDFOR
         \ENDIF

      \ENDFOR
   \ENDFOR
   \STATE \textbf{return} $\Delta \theta = \sum_i \mathbf{g}_i^{\text{PC}}$
\end{algorithmic}
\end{algorithm}
We show the full algorithm of Modular Gradient Surgery in Algorithm~\ref{alg:pcgrad}. Global Gradient Surgery is also shown in the algorithm.
\clearpage
\section{Additional Results on \twostage Training}
\label{sec:add_seq_results}
We provide the full results on \textit{Math} and \textit{Chat} for \twostage training checkpoints in Table~\ref{tab:full_results_sequential}.

\definecolor{graybg}{RGB}{245,245,245}
\definecolor{headerbg}{RGB}{230,230,240}

\begin{table*}[ht]
\centering
\caption{\textbf{Analysis of Sequential RL Training across Different Epochs and Orders.} We evaluate the trade-offs between Math and Chat capabilities. Math Avg. and Chat Avg. represent the mean performance across their respective benchmarks. All experiments use the Qwen-2.5-7B backbone.}
\label{tab:full_results_sequential}

\setlength{\tabcolsep}{4.5pt} 
\renewcommand{\arraystretch}{1.2}
\small

\begin{tabular}{ll cccc cccc}
\toprule

& & \multicolumn{4}{c}{\textbf{Math (avg@4)}} & \multicolumn{4}{c}{\textbf{Chat}}\\
\cmidrule(lr){3-6} \cmidrule(lr){7-10}

\rowcolor{headerbg}
\textbf{Strategy} & \textbf{Training Configuration} & \textbf{MATH} & \textbf{GSM} & \textbf{AIME} & \textbf{M. Avg.} & \textbf{WB} & \textbf{AE2} & \textbf{CW3} & \textbf{C. Avg.} \\
\midrule

\multicolumn{10}{c}{\cellcolor{graybg}\textit{Single-Domain Experts}} \\
\midrule
\multirow{2}{*}{Math-only} 
 & 2 Epochs & 77.2 & 91.8 & 15.0 & 61.3 & -27.1 & 4.4 & 32.7 & 3.3 \\
 & 4 Epochs & 77.1 & 92.1 & 11.7 & 60.3 & -14.1 & 7.5 & 30.9 & 8.1 \\
\arrayrulecolor{black!10}\midrule\arrayrulecolor{black}
\multirow{2}{*}{Chat-only} 
 & 2 Epochs & 71.8 & 82.7 & 10.0 & 54.8 & 44.7 & 57.2 & 41.7 & 47.9 \\
 & 4 Epochs & 72.7 & 86.8 & 16.7 & 58.7 & 46.9 & 57.5 & 47.6 & 50.7 \\
\midrule

\multicolumn{10}{c}{\cellcolor{graybg}\textit{\twostage RL}} \\
\midrule
\multirow{2}{*}{Math $\rightarrow$ Chat} 
 & Math$^{\text{2ep}}$ $\rightarrow$ Chat$^{\text{2ep}}$ & 75.9 & 90.8 & 13.3 & 60.0 & 35.7 & 35.4 & 43.9 & 38.3 \\
 & Math$^{\text{4ep}}$ $\rightarrow$ Chat$^{\text{2ep}}$ & 78.0 & 92.3 & 12.5 & 60.9 & 33.1 & 41.3 & 43.5 & 39.3 \\
\arrayrulecolor{black!10}\midrule\arrayrulecolor{black}
\multirow{3}{*}{Chat $\rightarrow$ Math} 
 & Chat$^{\text{2ep}}$ $\rightarrow$ Math$^{\text{2ep}}$ & 78.2 & 92.2 & 10.8 & 60.4 & 36.0 & 46.8 & 40.1 & 41.0 \\
 & Chat$^{\text{4ep}}$ $\rightarrow$ Math$^{\text{2ep}}$ & 77.6 & 91.5 & 15.8 & 61.6 & 52.0 & 49.3 & 40.4 & 47.2 \\
 & Chat$^{\text{4ep}}$ $\rightarrow$ Math$^{\text{4ep}}$ & 76.6 & 91.7 & 11.7 & 60.0 & 37.8 & 46.6 & 36.4 & 40.3 \\

\bottomrule
\end{tabular}
\end{table*}
\section{Additional Results on \singlestage Training}
\label{sec:additional_mgs}
\begin{figure}[ht]
    \centering
    \includegraphics[width=0.45\linewidth]{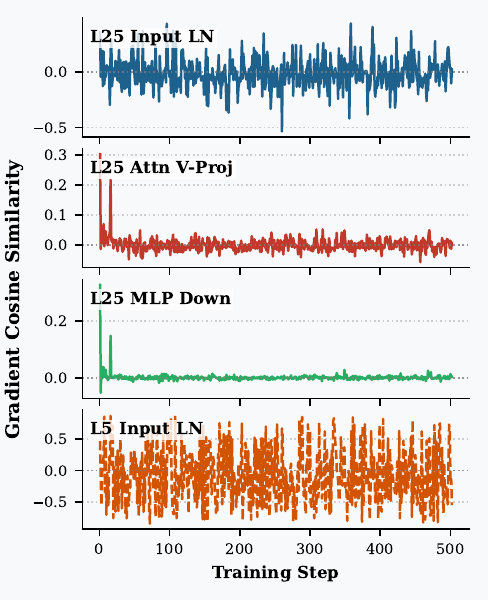}
    \caption{\textbf{Modular} gradient cosine similarity between \Math and \Chat in \texttt{Qwen-2.5-7B} during training. Gradient conflicts exist between different tasks across different training steps and modules.}
    \label{fig:grad_conflict_modular}
\end{figure}

\paragraph{Complementary Analysis: Modular gradient conflicts.}
Complementary to gradient conflict analysis shown in \ref{fig:grad_conflict}, we also show modular gradient conflicts in Figure~\ref{fig:grad_conflict_modular}.
\paragraph{Additional Analysis: Parallel training amortizes extra memory cost.}
Gradient surgery methods were known for their high memory overhead. However, this overhead can be significantly amortized in parallel training. Here we analyze the extra cost on time and memory, brought by MGS. For memory, the overhead is $\mathcal{O}(T\times N)$, where $T$ is the number of tasks and $N$ is the number of parameters. We need to store all the gradients and apply PCGrad before updating the model parameters. For time, the theoretical overhead is $\mathcal{O}(M\cdot T^{2})$, where $M$ is the number of modules when using \Module PCGrad, other $M=1$, and $T$ is the number of tasks. In practice, the time cost in applying PCGrad is $\sim163$ seconds in a $\sim30$ hours training, which is negligible. As for GPU memory overhead, the peak memory cost is $Memory_{\text{peak}}\approx(T_{\text{buffer}}+T_{\text{projected}})\times N \times Bytes$. However, due to the use of Fully-Sharded Data Parallel training, the actual cost for a single GPU is $\mathcal{O}(T\times \frac{N}{world\_size})$. Thereby, the peak memory overhead for a single GPU is $Memory_{\text{peak}}\approx(T_{\text{buffer}}+T_{\text{projected}})\times \frac{N}{world\_size} \times Bytes$, where $Bytes$ is determined by the training precision.
\clearpage
\section{Additional Results for Extended Experiments}
Here we show the concrete results of mixing three tasks and prolonged training. Results of mixing three tasks are in Table~\ref{tab:math_chat_if}. Results of prolonged training are in Figure~\ref{fig:scale_training_steps}.

As shown in \ref{tab:math_chat_if}, for \texttt{Qwen-2.5-7B}, \twostage training suffers from Mode Interference while \singlestage methods maintain high performance compared to their corresponding two-task counterparts.

\begin{figure}[ht]
    \centering
    \includegraphics[width=0.5\linewidth]{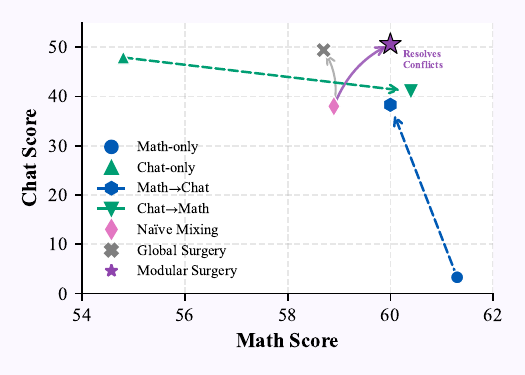}
    \caption{Performance of MGS compared with other methods. While both Modular and Global Gradient Surgery have higher \Chat score compared to Naïve Mixing, GGS is lower than MGS and has lower \Math score. MGS also achieves the highest \Chat score and surpasses Chat-only.}
    \label{fig:pareto_pcgrad}
\end{figure}












\begin{table}[h]
\centering
\caption{Results on mixing three tasks: \Math, \Chat, and \textit{IF}.}
\label{tab:math_chat_if}

\renewcommand{\arraystretch}{1.1}

\begin{tabular}{l ccccc}
\toprule
\rowcolor{headerbg}
\textbf{Method} & \textbf{Math} & \textbf{Chat} & \textbf{IF} & \textbf{Gen} & \textbf{Avg} \\
\midrule

\multicolumn{6}{l}{\texttt{\textbf{Qwen-2.5-7B}}} \\
\midrule
\multicolumn{6}{c}{\cellcolor{graybg} \twostage Training} \\
Chat-Math-IF & \underline{59.9} & 31.5 & \textbf{39.2} & 30.4 & 40.3 \\
Math-Chat-IF & 0.9 & 20.3 & 26.2 & \underline{35.3} & 20.2 \\
\addlinespace[3pt]
\multicolumn{6}{c}{\cellcolor{graybg} \singlestage Training} \\
Normal Mixing & 59.3 & \textbf{49.3} & 33.2 & 32.8 & \underline{44.6} \\
Global Surgery & \textbf{60.8} & 44.2 & 29.9 & 34.9 & 43.0 \\
\textbf{Mod. Surgery} & 58.8 & \underline{48.2} & \underline{33.7} & \textbf{35.7} & \textbf{45.0} \\ 

\midrule
\midrule

\multicolumn{6}{l}{\texttt{\textbf{Llama-3.1-8B}}} \\
\midrule
\addlinespace[3pt]
\multicolumn{6}{c}{\cellcolor{graybg} \singlestage Training} \\
Normal Mixing  & 35.0 & 22.9 & 25.4 & 25.2 & 27.3 \\
Global Surgery & \underline{35.8} & \underline{24.2} & \textbf{31.1} & \underline{30.3} & \underline{30.3} \\
\textbf{Mod. Surgery} & \textbf{36.0} & \textbf{30.5} & \underline{30.0} & \textbf{33.1} & \textbf{32.6} \\ 

\bottomrule
\end{tabular}
\end{table}

\clearpage
\section{Details of Single-task Baselines}
\label{sec:details_single_task_baselines}
Here, we include details of the single-task baselines:
\begin{enumerate}
    \item RLMT~\cite{bhaskar2025language}: this is a single-task baseline that leverages a reward model to train a reasoning LLM with thinking. In our setting, it is equivalent to Chat-only.
    \item OpenThinker v3~\cite{guha2025openthoughts}: this is an expert model in \Math with multi-stage training.
    \item General Reasoner~\cite{ma2025generalreasoneradvancingllmreasoning}: a model that trained on heterogeneous data with a unified reward. Despite the data diversity, we consider it as a single-task baseline because all the data they trained on isn't open-ended and has verifiable answers.
\end{enumerate}

\clearpage
\section{Case Study}
Here we demonstrate two cases from the checkpoint of \singlestage RL with MGS, trained on \Math, \Chat and \textit{IF}. One example is from MATH500 and another from Alpaca. As we can see, the thinking procedures do not contain clear step-by-step structured planning, which is different from the thinking procedures reported in previous single-task checkpoints.
\definecolor{boxbg}{HTML}{4527A0}
\begin{tcolorbox}[
    colframe=boxbg,                 
    colback=white,                  
    coltitle=white,                 
    title={Case Study: MATH500},        
    fonttitle=\bfseries\rmfamily,   
    arc=1mm,                        
    boxrule=1.5pt,                  
]
    \rmfamily 
    \textbf{Question:} \\
    Convert the point $(0,3)$ in rectangular coordinates to polar coordinates. Enter your answer in the form $(r,\theta),$ where $r > 0$ and $0 \le \theta < 2 \pi.$

    \vspace{1em}
    \textbf{Answer:} \\
    \texttt{<think>} \\
    To convert the point (0, 3) from rectangular (Cartesian) coordinates to polar coordinates, we need to determine the values of r and $\theta$. In polar coordinates, a point is represented as (r, $\theta$), where:
    \begin{itemize}
        \item r is the radial distance from the origin (0,0) to the point.
        \item $\theta$ is the angle measured from the positive x-axis to the line connecting the origin and the point.
    \end{itemize}

    The formulas to convert from rectangular to polar coordinates are: \\
    $r = \sqrt{x^2 + y^2}$ \\
    $\theta = \arctan(y / x)$ if $x \neq 0$, otherwise $\theta = \pi/2$ if $y > 0$, $3\pi/2$ if $y < 0$, and undefined if $y = 0$ and $x = 0$ (which is not the case here).

    For the given point (0, 3): \\
    $x = 0$ and $y = 3$.\\
    \texttt{</think>}

    \vspace{1em}
    \texttt{<response>} \\
    First, calculate r (the radial distance):
    \[ r = \sqrt{x^2 + y^2} = \sqrt{0^2 + 3^2} = \sqrt{9} = 3 \]

    Second, calculate $\theta$ (the angle): \\
    Since $x = 0$ and $y > 0$, the point (0, 3) lies on the positive y-axis. The angle $\theta$ from the positive x-axis to the positive y-axis is $\pi/2$ radians.
    \[ \theta = \frac{\pi}{2} \]

    Therefore, the polar coordinates of the point (0, 3) are (3, $\pi/2$).

    The answer within the required format is:
    \[ \boxed{\left(3, \frac{\pi}{2}\right)} \]
    \texttt{</response>}
\end{tcolorbox}
\definecolor{boxbg}{HTML}{4527A0}

\begin{tcolorbox}[
    colframe=boxbg,
    colback=white,
    coltitle=white,
    title={Case Study: Alpaca},
    fonttitle=\bfseries\rmfamily,
    arc=1mm,
    boxrule=1.5pt,
    breakable, 
    enhanced   
]
    \rmfamily 
    \textbf{Question:} \\
    What are the names of some famous actors that started their careers on Broadway?

    \vspace{1em}
    \textbf{Answer:} \\
    \texttt{<think>} \\
    To reason about this query, I will think about notable actors who started their careers on Broadway and have gained fame beyond the stage. I'll consider the most well-known Broadway productions, actors who have won major awards, and those who have transitioned to film, television, and other forms of media. I'll also consider recent trends and popular actors to make sure the list is relevant and up-to-date. I'll aim to provide a diverse selection of names to represent various eras and genres in Broadway history.\\
    \texttt{</think>}

    \vspace{1em}
    \texttt{<response>} \\
    Here are some famous actors who started their careers on Broadway and have achieved significant fame in film, television, and other forms of media:

    \begin{enumerate}
        \item \textbf{Meryl Streep}
        \begin{itemize}
            \item \textit{Broadway Debut}: "The Main Stem" (1970), "The Lion in Winter" (1980)
            \item \textit{Achievements}: Considered one of the greatest actresses of all time, she has won three Oscars, four Golden Globes, and an Emmy. Her Broadway career started in the early 1970s, and she has continued to be a leading figure in film and television.
        \end{itemize}

        \item \textbf{Daniel Day-Lewis}
        \begin{itemize}
            \item \textit{Broadway Debut}: "The Miracle Worker" (1979)
            \item \textit{Achievements}: Won three Oscars, a Golden Globe, and a BAFTA. His Broadway debut role as Anne Sullivan in "The Miracle Worker" was widely praised and marked the beginning of his international acting career that spanned decades.
        \end{itemize}

        \item \textbf{Brendan Fraser}
        \begin{itemize}
            \item \textit{Broadway Debut}: "The Boys from Syracuse" (1987)
            \item \textit{Achievements}: Known for his roles in "The Mummy" series, "The Great Muppet Caper," and "Mary Shelley's Frankenstein." He began his career on Broadway and transitioned successfully to film and television.
        \end{itemize}

        \item \textbf{Hugh Jackman}
        \begin{itemize}
            \item \textit{Broadway Debut}: "Les Misérables" (1991)
            \item \textit{Achievements}: Won the Tony Award for Best Actor in a Leading Role for "Les Misérables" and became internationally famous for his roles in the X-Men film series, "The Greatest Showman," and television roles like "How to Get Away with Murder."
        \end{itemize}

        \item \textbf{Julia Roberts}
        \begin{itemize}
            \item \textit{Broadway Debut}: "Steel Magnolias" (1988)
            \item \textit{Achievements}: Won an Oscar for "Erin Brockovich" and has been nominated for several others. Her Broadway debut was highly successful, and she has had a prolific career in films like "Pretty Woman," "Notting Hill," and "Eat Pray Love."
        \end{itemize}

        \item \textbf{Maggie Gyllenhaal}
        \begin{itemize}
            \item \textit{Broadway Debut}: "Love! Valour! Compassion!" (2000)
            \item \textit{Achievements}: Has won an Academy Award ("Crazy Heart"), a Golden Globe, and a Tony Award ("Love! Valour! Compassion!"). Her Broadway debut was a turning point in her career, and she has appeared in numerous acclaimed films and television shows.
        \end{itemize}

        \item \textbf{Ewan McGregor}
        \begin{itemize}
            \item \textit{Broadway Debut}: "The Rocky Horror Show" (1993)
            \item \textit{Achievements}: Known for his roles in films such as "Trainspotting," "Star Wars: Episode II – Attack of the Clones," and "The Last King of Scotland." His Broadway debut was memorable, and he has maintained success in both theater and film.
        \end{itemize}

        \item \textbf{Al Pacino}
        \begin{itemize}
            \item \textit{Broadway Debut}: "Dogfight" (1973)
            \item \textit{Achievements}: Won two Oscars ("Serpico" and "The Godfather"), a Tony Award ("Dogfight"), and numerous other accolades. His Broadway debut was early in his career and set the stage for his legendary film career.
        \end{itemize}

        \item \textbf{Viola Davis}
        \begin{itemize}
            \item \textit{Broadway Debut}: "King Hedley II" (2001)
            \item \textit{Achievements}: Won an Oscar for "Fences," a Tony Award ("Doubt"), and has been nominated for multiple Emmys. Her Broadway debut was critically acclaimed and launched her into major film roles and TV stardom.
        \end{itemize}

        \item \textbf{Christopher Walken}
        \begin{itemize}
            \item \textit{Broadway Debut}: "The Electric Company" (1972)
            \item \textit{Achievements}: Has had a long and successful film and television career with roles in iconic films like "The Deer Hunter," "Killer Klowns from Outer Space," and "Ghostbusters." His Broadway debut was in a prominent Off-Broadway play but helped establish him as an actor of depth and versatility.
        \end{itemize}
    \end{enumerate}

    \textbf{Reasoning Behind Selection:}
    \begin{itemize}
        \item \textbf{Notoriety and Awards}: These actors have all achieved significant recognition, including Academy Awards, Golden Globes, Emmy Awards, and Tony Awards.
        \item \textbf{Broad Range of Careers}: Transition across film, TV, and voice acting.
        \item \textbf{Diversity in Roles and Styles}: From comedy to musicals and heavy drama.
        \item \textbf{Recent Success}: Inclusion of active stars to ensure contemporary relevance.
        \item \textbf{Representative of Different Eras}: Spanning decades of Broadway history.
        \item \textbf{Cultural Impact}: Household names defining popular culture.
    \end{itemize}

    \textbf{Additional Considerations:}
    \begin{itemize}
        \item \textbf{Pop Culture Relevance}: Stars like Hugh Jackman returning for "The Music Man."
        \item \textbf{Variety}: Mix of nationalities and genres.
    \end{itemize}

    \textbf{Why Others Might Not Be Listed:}
    \begin{itemize}
        \item Denzel Washington and Tommy Lee Jones had Broadway stints but are more strictly identified with film/TV.
    \end{itemize}

    \textbf{Conclusion:}
    These actors are famous for starting their careers on Broadway and going on to have hugely successful careers in film and television. Their experiences on Broadway undoubtedly contributed to their ability to portray complex characters and connect with audiences across multiple media.

    \vspace{0.5em}
    \textbf{Additional Names:}
    \begin{itemize}
        \item Denzel Washington, Tom Hanks, Laurence Fishburne, Kathryn Hahn, Jeffrey Wright.
    \end{itemize}

    \textbf{Significance:}
    \begin{itemize}
        \item \textbf{Longevity}: Remaining active in theater (e.g., Jeffrey Wright).
        \item \textbf{Critical Acclaim}: Theater work often precedes film recognition.
        \item \textbf{Versatility}: Transferring stage skillsets to the screen.
    \end{itemize}

    \textbf{Final Thoughts:}
    Broadway remains a vital launching pad for Hollywood talent. The challenging environment of live performance helps shape the craft and confidence required to succeed in global entertainment.\\
    \texttt{</response>}
\end{tcolorbox}

\section{Model Template}
Because our work focuses on RL training from a base models without a preceding Supervised Fine-tuning (SFT) phase, aka., Zero-RL, we use a customized chat template to incentivize a CoT process. The template is:

\definecolor{boxbg}{HTML}{4527A0}
\begin{tcolorbox}[
    colframe=boxbg,                 
    colback=white,                  
    coltitle=white,                 
    title={Chat Template},    
    fonttitle=\bfseries\rmfamily,   
    arc=1mm,                        
    boxrule=1.5pt,                  
]
    \rmfamily 
    A conversation between User and Assistant. Following the User’s query, the Assistant first plans a response, and then provides the response. The internal reasoning process is enclosed within \texttt{<think>} \texttt{</think>} tags and the response is enclosed within \texttt{<response>} \texttt{</response>} tags, i.e., in the format \texttt{<think>} reasoning process here \texttt{</think>} \texttt{<response>} response here \texttt{</response>}. \\
    User: \texttt{<query>} ...user text... \texttt{</query>} Assistant:
\end{tcolorbox}
\section{Complete Related Works}
\label{sec:complete_related_works}
\paragraph{LLM RL Post-training}
LLM post-training typically refers to the sequence of alignment and specialization stages applied to a base model before deployment. Reinforcement learning has become a central component of this process, most notably through reinforcement learning with human feedback (RLHF)~\citep{christiano2017deep,ouyang2022training,bai2022training} and more recently reinforcement learning with verifiable rewards (RLVR). RLVR has been shown to significantly improve reasoning performance in domains with objective correctness signals such as mathematics and coding~\citep{wen2025reinforcement,zeng2025simplerl,shi2025deep}.
Other works such as rubric-based~\cite{liu2025openrubrics,gunjal2025rubrics,dineen-etal-2025-qa} and checklist-based~\cite{viswanathan2025checklists} rewards show the possibility of LLM RL training with miscellaneous reward functions~\cite{chang2025bleuberi,zhang2025tdrm,guo2025reward}.
Recent work demonstrates that RL can also improve open-ended chat and instruction following. \citeauthor{bhaskar2025language} show that applying RL on top of reasoning traces leads to stronger conversational ability, while \citeauthor{zeng2025zero} demonstrate that zero-RL can generalize beyond strictly verifiable domains. Large-scale systems such as DeepSeek-R1 \citep{Guo2025DeepSeekR1} further validate the effectiveness of RL for reasoning, though they rely on complex multi-stage pipelines involving cold-start SFT~\cite{seed2025seed1}, rejection sampling~\cite{xiong2025minimalist}, and multiple rounds of RL.
Despite these advances, existing RLHF and RLVR approaches typically focus on a single target domain at a time. How to jointly train a model that is strong at both verifiable reasoning (e.g., Math) and open-ended interaction (e.g., Chat) using a simple RL recipe remains under-explored. Our work addresses this gap by studying RL post-training with multiple heterogeneous rewards simultaneously.

\paragraph{General Multi-task RL}
Our \singlestage training setting is closely related to classical multi-task reinforcement learning, where a single policy is optimized over multiple tasks. A central challenge in this literature is negative interference between tasks, especially when gradients are imbalanced.
A line of work addresses this issue through Gradient Manipulation methods~\cite{chen2025gradient}, including GradVec \citep{wang2020gradient}, PCGrad \citep{yu2020gradient}, and GradDrop \citep{chen2020just}, which aim to find Pareto-improving or conflict-averse updates.
Another line of work addresses gradient imbalance through Gradient Weighting. This line of work aims at finding a common update direction by adaptively aggregating gradients from different objectives/tasks~\cite{kim2025conflictaverse,wang2025soft,liu2021conflict,sener2018multi,chen2018gradnorm}.
PCGrad, in particular, performs gradient surgery by projecting conflicting task gradients to remove destructive components, and has been shown effective in both supervised learning and RL settings. However, these methods treat model parameters as a monolithic vector.
In contrast, large language models exhibit strong modular structure across attention, MLP, and normalization components. Our work extends classical gradient conflict resolution to LLM RL by introducing Modular Gradient Surgery, which resolves conflicts at the module level rather than globally. This design better aligns with transformer architectures and leads to improved stability and performance in multi-domain RL post-training.

\paragraph{LLM Multi-task RL}
Several recent works explore multi-task or multi-objective RL in the context of large language models~\citep{feng2026rlar,zhang2025agentrl,phan2025beyond}. \citeauthor{Guo2025DeepSeekR1} perform mixed RL training across Math, STEM, and general domains, showing that multi-task zero-RL can be effective with appropriate regularization. OmniThinker \cite{li2025omni} scales multi-task RL using hybrid rewards and task scheduling strategies. Industrial systems such as DeepSeek-V3.2 \cite{liu2025deepseek} report mixed RL training after expert distillation, though detailed analyses are not publicly available. MiMo-VL-Miloco~\cite{li2025xiaomi} from Xiaomi also conducts mixed training for tasks from different modality.
Apart from LLM reasoning, LLM multi-task learning is more often seen in tasks such as LLM unlearning~\cite{jin2025unlearning}, whose innate goal is to optimize two conflicted objectives. Other tasks such as LLM safety~\cite{yi2025gradient} may also aim at optimizing multiple conflicting goals at the same time.
However, in LLM RL post-training, few tryouts have been taken to address gradient imbalance. \citeauthor{wu2025imbalanced} is the first to identify the significance of gradient imbalance in LLM RL post-training, and study through the perspective of Gradient Weighting. They find that the imbalance can hardly be explained by a series of metrics, e.g., learning gains, advantages, etc. Our work serves as the first to address this imbalance through Gradient Manipulation and showcase that resolving the conflicts/imbalances in this way leads to substantial and consistent gains.


\end{document}